*Perspective on Utilizing Foundation Models for Laboratory Automation in Materials Research*

Authors

Kan Hatakeyama-Sato*[1,2], Toshihiko Nishida[3], Kenta Kitamura[4], Yoshitaka Ushiku,[2,5,6,7] Koichi Takahashi,[2,7] Yuta Nabae[8], and Teruaki Hayakawa[8]

1 Department of Technology Management for Innovation, The University of Tokyo., Japan.

2 Laboratory for Biologically Inspired Computing, RIKEN Center for Biosystems Dynamics Research., Japan

3 Department of Chemistry, The University of Tokyo*,* Tokyo 113-0033, Japan

4 Intellectual Moonshine, Japan.

5 OMRON SINIC X Corporation, Tokyo, Japan

6 NexaScience, Tokyo, Japan

7 RIKEN Advanced General Intelligence for Science Program, Japan

8 Materials Science and Engineering, School of Materials and Chemical Technology, Institute of Science Tokyo. Tokyo 152-8552, Japan

E-mail: kan.hatakeyama [[at]] weblab.t.u-tokyo.ac.jp

**Abstract**

This review explores the potential of foundation models to advance laboratory automation in the materials and chemical sciences. It emphasizes the dual roles of these models: cognitive functions for experimental planning and data analysis, and physical functions for hardware operations. While traditional laboratory automation has relied heavily on specialized, rigid systems, foundation models offer adaptability through their general-purpose intelligence and multimodal capabilities. Recent advancements have demonstrated the feasibility of using large language models (LLMs) and multimodal robotic systems to handle complex and dynamic laboratory tasks. However, significant challenges remain, including precision manipulation of hardware, integration of multimodal data, and ensuring operational safety. This paper outlines a roadmap highlighting future directions, advocating for close interdisciplinary collaboration, benchmark establishment, and strategic human-AI integration to realize fully autonomous experimental laboratories.

## 1. Expectations for Foundation Models in Materials Laboratory Automation

Laboratory automation, a technology aimed at automating experimental research, is expected to pave the way for a new research paradigm in materials science [1, 2, 3]. By rapidly and comprehensively executing numerous experiments, laboratory automation accelerates research, enhances reproducibility through precisely controlled robotic processes, and enables swift and distributed knowledge sharing among researchers worldwide [1]. This technology is anticipated to contribute significantly to the development of crucial devices and compounds, including catalysts for energy and chemical conversions, environmentally friendly plastics, solar cells, secondary batteries, fuel cells, thermoelectric conversion modules, nuclear fusion reactors, quantum computers, and energy-efficient computing systems [1, 4, 5].

The success of next-generation laboratory automation depends not only on experimental hardware but also on the utilization of artificial intelligence (AI), especially foundation models. Foundation models represent a new AI paradigm encompassing large language models like GPT-4 [6], multimodal models, and agent-related technologies. These foundation models and generative AI have begun to influence chemistry and materials science [7], giving rise to diverse applications including molecular and materials design [8, 9, 10], reaction pathway exploration [11], catalyst design [12], and even autonomous planning of chemical experiments [13]. Additionally, foundation models are being expanded to hardware control mechanisms, enabling natural language-driven robotic operations [14, 15].

In natural sciences, AI can help analyze high-dimensional spaces and complex interactions that are challenging to address using conventional human cognition [16, 17]. For instance, the Nobel Prize-winning AlphaFold in 2024 [18, 19] demonstrated that deep learning significantly outperforms traditional methods in predicting protein folding structures. The complex interplay of factors contributing to protein folding structures is difficult for humans to precisely grasp. In contrast, the transformer algorithms embedded in the AI model can work beyond human cognitive constraints, allowing predictions of multi-system interactions in high-dimensional vector spaces from extensive datasets [18, 19]. By integrating AI capabilities—characterized by their strength in handling complexity and broad cognitive abilities across various disciplines [16] —with laboratory automation, it is anticipated that new materials can be efficiently discovered. This is made possible by AI capacity to learn and analyze complex correlations among molecular structures, processes, and properties [13, 20]. Furthermore, the integration of foundation models with robotics will enable complex experimental manipulations that specialized automated experimental systems previously found challenging.

The technical roles of foundation models in materials laboratory automation can be broadly categorized into (1) models that predict chemical and materials phenomena for experimental planning and (2) models that control experimental devices and robotic handlers [1]. The first category has been partially explored in materials informatics and cheminformatics, particularly for studying structure-property correlations [21, 22]. Introducing foundation models capable of intricately combining different concepts is expected to enhance multi-scale simulations, flexible integration with real-world data, and comprehensive collaboration with process and device sciences, leading to more accurate predictions [1, 13, 20]. In contrast to traditional informatics approaches that require separate constructed predictive models and databases for each task or project, foundation models potentially integrate all scientific knowledge, revealing hidden interdisciplinary commonalities difficult for humans to uncover. Beyond experimental result predictions, foundation models hold the potential to handle a broad spectrum of intellectual tasks, including research theme exploration, experiment planning, scheduling, report preparation, and peer review [23, 24]. The second category, robotics-oriented foundation models, intersects AI and robotics, with initial applications being explored in factory settings and home tasks like cooking. However, foundation model development specifically for the experimental sciences remains largely unexplored.

In this review, we focus on the emerging AI paradigm of foundation models, outlining recent advances and challenges in AI-robot integrated laboratory automation and predicting prospects. We particularly examine the potential for developing laboratory automation technologies capable of handling open-ended environments. Conventionally, research has been carried out in closed, standardized environments designed to facilitate data, equipment, and communication formats amenable to AI and robotics [1]. While this standardized approach suits software and hardware with limited capabilities and simplifies the rule-based connections of substances and information across various devices, it risks compromising the intrinsic flexibility, convenience, and scalability of experimental research. As a countermeasure, combining foundation models with general-purpose robots is gaining attention as a promising technology capable of autonomously conducting diverse experimental studies. Thus, this review introduces recent trends in foundation models and robotics related to laboratory automation. The roles of foundation models can be broadly divided into two domains: cognitive (brain) and physical (body) (Figure 1). Cognitive roles encompass intellectual tasks such as designing research plans (including literature surveys), formulating experimental plans, processing data, and writing reports. Physical tasks, essential in real-world experiments, include controlling experimental hardware, sensing, and orchestrating multiple instruments.

For clarity, this review primarily focused on practical laboratory experiments, excluding detailed discussions on structure-property correlation predictions [22], virtual chemical structure generation, simulation utilization, and ethical issues [9, 11, 25]. These topics require further exploration in existing reviews [1, 2, 9, 11, 25, 26] or future comprehensive analyses.

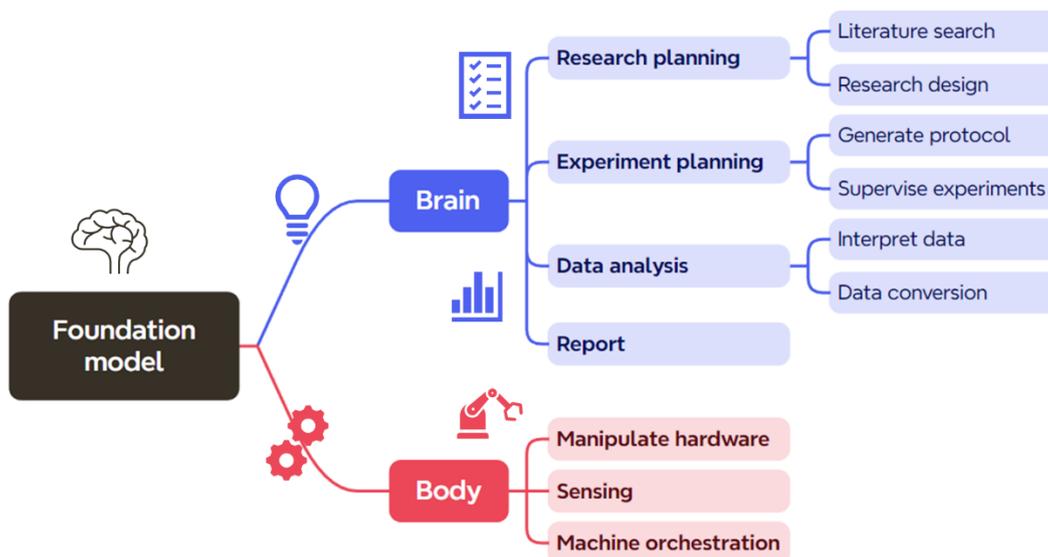

**Figure 1** Major roles of foundation models for laboratory automation.

## 2. Overview of Current Lab Automation

### 2.1 Specialized Hardware Systems and Module Standardization

Throughout the history of automation in material and chemical synthesis, specialized hardware systems have been actively developed to efficiently perform specific experimental tasks [1, 27]. Representative examples include automated experimental systems designed to conduct pre-defined synthesis reactions and high-throughput screenings with high precision, speed, and automated operations such as sample preparation, reaction control, and measurement. This trend began with the development of peptide automated synthesizers in the 1960s, followed by the commercialization of laboratory robots in the 1980s. During the 1990s, high-throughput screening and combinatorial chemistry became widespread, significantly enhancing the experimental efficiency [28]. Parallel reactors and automated synthesizers expanded rapidly in the 2000s, accelerating the development of efficient catalysts and materials [28]. By the 2010s, general-purpose automated synthesizers and systems like Chemputer gained attention, enabling comprehensive automation from synthesis to evaluation (Figure 2a) [29, 30].

More recently, systems integrating advanced algorithms such as Bayesian optimization and reinforcement learning have emerged, enabling the optimization of material compositions and reaction conditions. Additionally, autonomous robots that can navigate laboratories and conduct experiments independently have been developed, bringing fully autonomous laboratories closer to reality [31, 32]. Applications have continued to expand, including high-throughput equipment for inorganic material synthesis [33] and automated battery performance evaluation systems [34]. Moving forward, standardizing and modularizing specialized hardware is essential for achieving more flexible and versatile lab automation systems. A leading example in Japan is the Digital Laboratory (dLab), a fully autonomous platform that has demonstrated the self-directed synthesis of $LiCoO_2$ thin films [35] (Figure 2b). dLab employs a standardized data format called MaiML and modular measurement and control units. Through close collaboration between academia and industry, it is advancing both the extension of this approach to ceramic materials and the broader standardization and modularization of laboratory hardware.

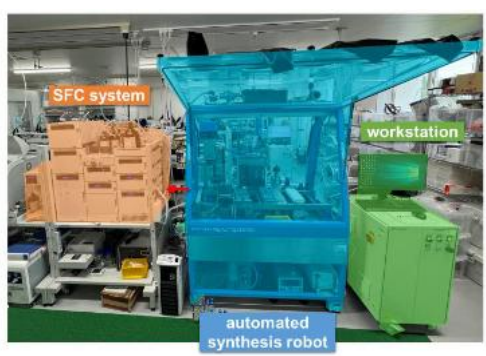

a)

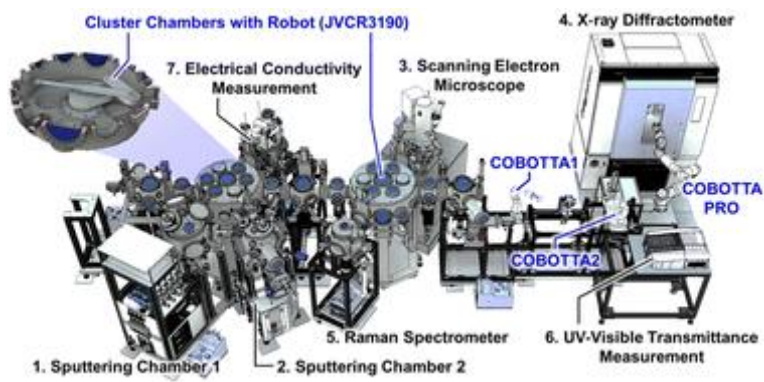

b)

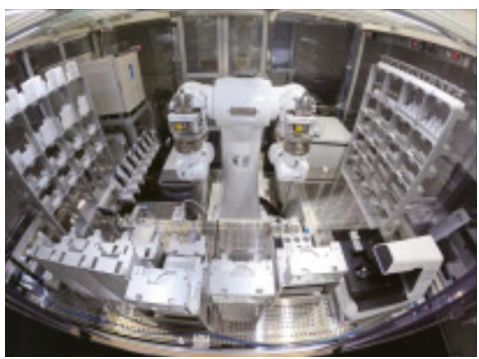

c)

**Figure 2**. Images of automated laboratory systems. a) Automatic autonomous optimization system for chemical reactions (green: a workstation); blue: robot; orange: supercritical fluid chromatography (SFC) system. The image was reprinted from a reference under the CC-BY-NC-4.0 license [32]. b) The automatic and autonomous optimization system for solid materials research (dLab). Image reprinted from the original reference under a CC BY-NC 4.0 license [35]. c) Robotic cell processing facility featuring Maholo LabDroid. The image was reprinted from a reference under the CC-BY-4.0 license [36].

Alongside chemistry and materials science, significant research and development in laboratory automation is advancing in the field of biomedical sciences (bio) [37, 38]. A representative example is the Robot Scientist developed by Ross D. King et al [39, 40]. In Japan, bio-research automation utilizing LabDroid "Mahoro" has been reported (Figure 2c) [36, 41, 42, 43]. Mahoro is distinguished by its design, which adapts industrial dual-arm robots for life science applications, capable of executing various protocols such as pipetting and cell culture [41]. Additionally, affordable dispensing devices designed for general scientists are commercially available and enable simple chemical experiments [13].

A challenge for specialized automated systems is their limited compatibility with experimental procedures, often restricted to specific tasks such as optimizing the composition ratios within fixed experimental protocols [1]. Conversely, in fundamental experimental research, procedures frequently change, making it challenging for specialized systems to adapt. Consequently, the extremely high cost of specialized systems is difficult to justify in many scientific research scenarios, hindering the broader adoption of laboratory automation. At the preliminary research level, there have been reports of handmade, low-cost implementations for tasks such as dispensing, polishing, and stirring mechanisms [26, 44]. While this approach offers advantages in terms of experimental flexibility and educational value, it faces issues regarding the human cost of device construction and maintenance, reproducibility, and inter-organizational sharing [1, 44].

To reduce the construction and updating costs of automated experimental systems, an approach that employs conventional hardware divided into standardized functional modules with clearly defined input-output interfaces is gaining attention [1]. By implementing tasks such as dispensing, weighing, grinding, and measurement as individual experimental modules, and connecting them through standardized communication protocols and data formats, it becomes possible to address a variety of experimental needs [45, 46]. Efforts to standardize data formats for experimental equipment have also been highlighted [46]. Standardizing experimental equipment and sample geometries, analogous to industrial manufacturing processes, is another important aspect of the modularization concept [47].

Scheduling optimization to facilitate smooth collaboration among modules is also a crucial element of laboratory automation. Recently, methods optimizing experimental protocols and task allocation across devices have demonstrated practicality in chemistry and life sciences [47, 48]. Additionally, the automation of operational management and troubleshooting has been proposed after the introduction of autonomous systems [49]. However, achieving comprehensive autonomy remains challenging due to mathematically formulating user requirements, which are often qualitative and difficult to quantify, as well as the necessity for flexible equipment switching [1].

## 3. Future Laboratory Automation without Standardization
### 3.1 Limitations of Standard Modularization and Foundation Models as a Next-Generation Approach

Modularization and standardization of automated experimental systems are anticipated to become major trends; however, it remains uncertain whether these approaches can cover all experimental research [1]. The costs associated with unifying the devices, measuring instruments, and communication standards cannot be ignored. Moreover, these strategies sometimes conflict with manufacturers' interests, who aim to retain users through proprietary standards, suggesting that perfect standardization might be difficult. Cutting-edge synthetic and analytical techniques often emerge through unique and pioneering devices before standardization takes place, and serendipitous discoveries frequently occur outside standardized frameworks.

An alternative to standardized laboratory automation is automating diverse experimental research and data processing currently performed by humans using foundation models [9, 11, 25]. By developing humanoid robots with advanced cognitive and operational capabilities, it may become possible for these robots to navigate laboratory environments and operate a wide range of equipment and instruments similarly or more skillfully than humans. In this scenario, robots could effectively leverage multimodal information such as vision, tactile sensations, and audio signals, even without uniformly modularizing various instruments and devices. Currently, autonomous robots have demonstrated effectiveness in laboratory settings, although many operational methods rely on rule-based methods that restrict their predefined tasks [31]. Achieving autonomous experimental robots capable of functioning in open-ended environments like humans will require next-generation hardware and software technologies.

The key technology for flexible adaptation to various sample types, devices, communication standards, and databases lies in highly versatile AI agents and foundation models for robots, which have gained significant attention in recent years [50, 51]. Versatile AI agents can autonomously evaluate situations with minimal instruction, completing desired tasks with practical accuracy comparable to humans [52]. Tasks such as research topic formulation, literature review, experimental planning, and data analysis are also within the capabilities of these AI agents [24]. Robotic foundation models serve as command centers for flexibly controlling interactive tasks in physical spaces through natural language interfaces similar to human communication [15, 51]. Even in stages where specialized devices and modules collaborate before humanoid robots become widespread, foundation models are considered valuable. They can automate the standardization of data obtained from analytical and measurement instruments and facilitate cooperation by integrating advanced AI technologies such as large language models (LLMs), robotic foundation models, and AI agents. Such flexible automation mechanisms enable efficient collaboration among modules and allow rapid responses to frequent changes in experimental protocols [9, 11, 25, 50, 51].

**3.2 Emergence of General AI Agents Leveraging Foundation Models**

With advancements in LLMs, a type of foundation model, practical applications of AI agents capable of sophisticated knowledge integration and autonomous language understanding have been actively explored since 2024 [53, 54]. As of February 2025, no systems have yet reached human-level information processing, but domain-specific models and services have begun to emerge. At the time of writing, major vendors of state-of-the-art AI models include OpenAI (GPT-4.5, o1, o3 series), Google (Gemini), and Anthropic (Claude). These frontier models have started to surpass expert-level performance in various scientific and technological examinations.

As of early 2025, the accuracy of the agents capable of general-purpose tasks remains limited, driving the active development of domain-specific agent systems. One area in which LLMs excel is programming, where performance surpassing that of average human programmers has been reported for certain tasks [55]. In laboratory automation, precise hardware control through programming is crucial. It is conceivable that within a few years, AI agents utilizing LLM-based reasoning could autonomously generate specific experiment-operation programs from experimental plans and perform operational fine-tuning independently [55].

Foundation models adapted to diverse data formats are also beginning to emerge. For example, AI agents specialized in databases have been developed [55, 56]. If database agents become capable of autonomously generating transformation code for structured data and further leverage LLMs to process linguistic nuances—such as standardizing unstructured data or resolving variations in notation—then the automatic generation of standardized data from various database formats could become partially feasible. However, adapting such models to scientific domains requires collaboration with experts in chemistry and materials science. Additionally, multimodal models capable of interpreting tables, graphs, and spectral data will be necessary [1]. Constructing AI agents that can autonomously handle comprehensive data-processing tasks is expected to significantly facilitate automatic data standardization and international data sharing.

**3.3 AI Agents for General Scientific Research**

Benchmarks and prototypes for AI agents designed to perform scientific research tasks such as proposing ideas, conducting experiments, analyzing data, and writing have started to appear [16, 57, 58, 59]. For example, an approach has been proposed for biomedical discovery, in which AI agents continuously learn within their specific domain and integrate scientific knowledge, biological principles, and theories using machine learning tools [60]. Fields such as computer science and mathematics, which do not require physical-world interactions, are actively exploring AI agents capable of automating scientific research and technological development. Frameworks are being developed to fully automate data science tasks related to text and image processing and other AI tasks, that have traditionally been research topics, competitions, or educational content [61, 62]. The "AI Scientist" prototype pipeline has gained attention as a system aimed at fully automating the scientific research process (Figure 3)[24]. Reported in the summer of 2024, this system claimed to automate research proposal generation, experimentation, manuscript writing, and even peer review completely. Impressively, the cost of conducting a single research project was estimated to be around a few tens of dollars, significantly cheaper than human-led research. Such cost advantages could significantly influence the competitive landscape against traditional human-centered research activities. Moreover, the fact that a single model learns across various scientific domains suggests that AI could become a valuable tool for discovering hidden interdisciplinary connections. Version 2, released in March 2025, marked the first time a paper authored by the AI Scientist was accepted into an international conference workshop [63].

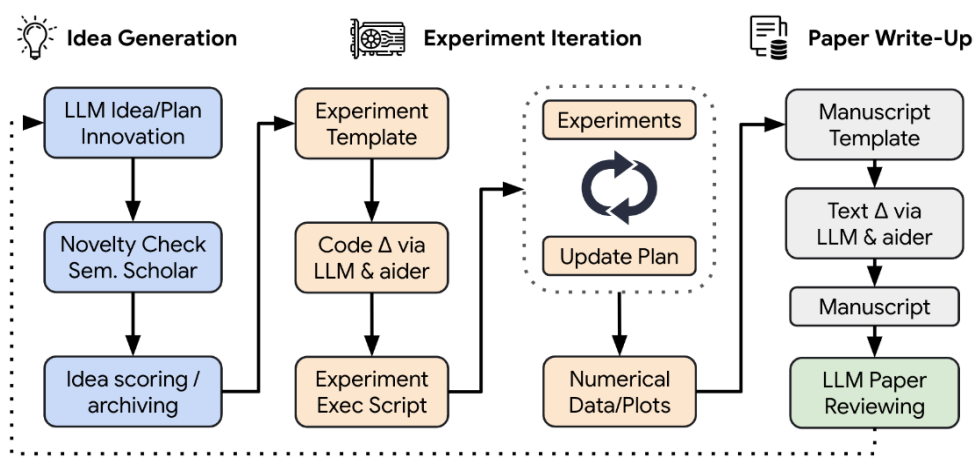

**Figure 3** Conceptual illustration of the AI scientist for fully automated research. The image was reprinted from a reference under the CC-BY-4.0 license [24].

The application of AI agents to research activities is promising; however, several limitations still need to be addressed. Key challenges include handling complex tasks, adapting to experimental scientific domains, and ensuring output reliability [57, 58, 59]. AI agents used for research and development require advanced capabilities for processing complex intellectual tasks. Notably, the number of trial-and-error attempts performed by AI in experiments is limited, typically fewer than those conducted in standard research activities [24]. This issue is partially due to the limited capability of LLMs to perform long-term tasks. To address this approaches such as equipping AI with human-like long-term memory are beginning to gain attention as potential solutions [64].At the same time, in chemistry and materials experiments, research based on simple combinatorial optimization, exemplified by Bayesian optimization, continues to yield significant results [31, 65]. This is largely due to the complexity of molecular systems and the difficulty in predicting experimental results, which suggest that extensive experimentation often correlates with successful outcomes. Thus, AI agents at the current development level could potentially achieve notable results in rapid experimental systems that rely more on trial and error than on deep thinking, such as optimizing conditions for specific systems.

AI agents for scientific research must meet various performance requirements beyond those mentioned previously. For instance, implementing open-close strategies in R&D demands AI systems to share general insights from confidential organizational information externally without exposing the confidential data itself. While secure computation research has been ongoing in the AI field, it remains an emerging area for LLMs [66]. Ensuring the scientific validity when AI agents report extensive research findings and analyses also poses a serious concern, necessitating discussion about the peer-review costs within the scientific community [24].

The significant constraints of the transformer architecture, predominantly used in current foundational models, must also be noted. The performance of transformer-based models is known to improve according to a scaling law as the amount of training data increases [67]. While this principle has enabled the development of high-performing LLMs, it also poses limitations on rapid performance improvements and knowledge retention. Humans can efficiently learn crucial information from just one or a few inputs, whereas transformers typically require hundreds of times more data for effective knowledge retention [68]. Literature describing cutting-edge scientific research often exists in limited forms, such as single original papers or isolated experimental notes. Consequently, a naive approach of merely training models on scientific data may not effectively capture user-expected information [69]. Expanding methodologies by understanding scaling laws related to expert knowledge acquisition and utilizing synthetic data for data augmentation will be crucial for efficiently imparting knowledge [68]. To overcome the foundational model limitations discussed here, cognitive augmentation technologies, particularly specialized external cognitive systems for scientific research (Science Exocortex), combining human researchers' intellectual abilities with AI, could emerge as a significant area of future research [70].

### 3.4 Foundational Models for Chemistry and Materials Experiments

Research utilizing foundational models as agents in chemistry and materials science is beginning to emerge, including reports on optimizing chemical reaction conditions [13], end-to-end foundational model applications [71], and frameworks for exploring physical chemistry laws [72]; however, such implementations remain limited. This limitation largely stems from the significantly lower versatility of automated experimental hardware compared to human capabilities [1]. Besides enhancing automated experimentation equipment, which remains a key implementation bottleneck, it is also essential to explore methods for equipping foundational models with specialized knowledge in chemistry and material science. Currently, the primary data sources for training widely available LLMs are internet texts, which often do not align with domains necessary for laboratory automation. While LLMs trained on chemical structures have been reported, much information arising within laboratories remains undocumented in scientific literature or general web content, leading to an emerging awareness that models trained exclusively on general information sources exhibit limited ability to accurately interpret experimental phenomena. A clear example is ChatGPT, which reportedly possesses Ph.D.-level chemistry knowledge, but sometimes misidentifies laboratory glassware, illustrating performance challenges in applying these models to experimental sciences (Figure 4).

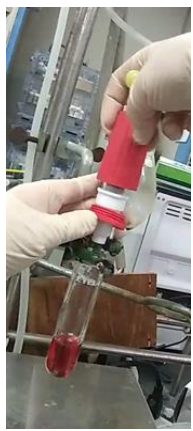

**Figure 4** Example of laboratory equipment image misrecognition by ChatGPT (GPT-4.5)**.** Incorrect recognition by GPT-4.5: "A pipette connected to a pipette pump transferring liquid into a glass test tube." Correct: "A glass tube equipped with a sealing valve designed for inert-atmosphere reactions" (inference conducted in April 2025 by the authors).

It remains unclear how well current foundational models perform experimental scientific tasks from a multimodal perspective, such as visual and auditory modalities. Moving forward, collaboration between experimental scientists and AI researchers will be essential for generating training datasets and quantitative benchmarks that enable models to recognize laboratory information effectively [73,

74]. These datasets and benchmarks must also include laboratory information not previously digitized thoroughly, such as details about the size and specifications of glassware used and nuanced experimental techniques and know-how.

Research into hardware and software designed that can analyze, digitize, and comprehensively transfer the insights of skilled practitioners to AI is also necessary. Besides standard methods like using first-person cameras to observe experiments, eye-tracking technology could be employed to record and predict the focal points of skilled technicians [75]. Efforts to sense, database, and model sensory information—such as hearing, smell, touch, and even infrared sensing—across diverse humans and other organisms, will become increasingly important [76].

Improving the reliability of model outputs is another critical research area, especially for scientific experiments involving real-world tasks, where ensuring safety becomes paramount. While programming-oriented agents currently moving towards practical applications are growing more reliable, they still occasionally make errors [2, 53]. AI issuing instructions for hazardous chemical reactions could endanger hardware and operators, so such situations must be avoided as thoroughly as possible. To guarantee reliability and safety, multiple combined approaches are essential. Primarily, it is necessary to continuously implement human-in-the-loop cycles—such as improving the model, monitoring and restricting experimental plans by humans, and providing feedback—to enhance output accuracy [77].

Constructing a safe, virtual sandbox environment is also important [57, 59, 61, 62]. Using a 3D physics simulator as a sandbox for virtual trial-and-error is significant in laboratory automation. Simulation in virtual spaces is common in robotics, where AI-controlling robots typically operate within virtual environments [78]. Simulators are also partially utilized in experimental science. For instance, using physical simulators to facilitate robot arms in weighing powders has been proposed in anticipation of material experiments [79]. In robotics, virtual environments leveraging 3D game engines such as Unity and Unreal Engine are increasingly available [80]. Efforts to generate virtual environments using foundational models (world models) are also beginning to gain momentum [81].

Research on constructing digital twins reflecting real-world characteristics is also beginning to emerge in laboratory automation (Figure 5) [82]. It is possible to experience experimental environments virtually through virtual reality (VR), with potential applications including the virtual design of automation centers and scientific education. Moving forward, constructing more advanced 3D simulation platforms capable of handling chemistry and material experiments likely become essential [82].

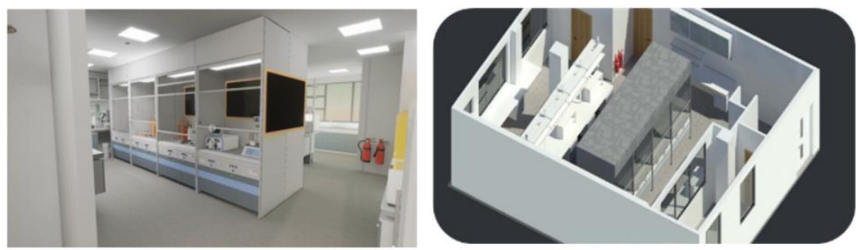

**Figure 5** 3D images of a digital twin laboratory using an Unrealistic engine. The images were reprinted from a reference under the CC-BY-4.0 license [82].

## 4. Robot Foundation Models

### 4.1 Research Trends and Challenges Before Foundation Models

Implementing laboratory automation in physical spaces requires robots—hardware designed to faithfully execute computer-generated experimental plans [1]. Previously, robot control often relied on rule-based systems or partially utilized specialized AI modules. Although rule-based approaches offer reliability, they struggle to handle unexpected phenomena frequently encountered in experiments and incur high costs for system development and updates. Therefore, systems combining the precision and reproducibility of robots with human-like flexibility have become highly desirable.

Recently, the robotics field has witnessed innovative progress with the introduction of large-scale general-purpose models, known as "Foundation Models" [[15, 51, 78, 83]. Traditional robot control AI typically involves task-specific or module-specific developments. However, general-purpose models pretrained on vast amounts of image, language, and action data have shown the potential to adapt across multiple tasks and handle unknown situations via zero-shot learning, without requiring task-specific training [15, 51, 78]. Such characteristics could significantly contribute to solving issues related to the cost and rigidity of control mechanisms in laboratory automation [83].

Historically, robot learning has relied heavily on reinforcement learning methods, where agents were individually trained from scratch for each specific task [83]. This approach required the design of task-specific reward functions and tuning hyperparameters for each application (such as specific robot operations or games), often involving extensive trial-and-error in both simulations and real-world environments. Consequently, it was challenging to transfer knowledge between tasks, limiting reuse and generalization [83]. Similar learning constraints are shared in fields such as materials informatics, cheminformatics, and laboratory automation, where researchers often have to develop specialized predictive, analytical, and proposal systems from scratch for each project, leading to poor compatibility between projects—a limitation inherent to traditional AI algorithms with restricted generalizability [11, 83, 84].

**4.2 Transformative Impact of Foundation Models**

Since the 2020s, foundation model technologies, particularly transformers, have increasingly been applied in robotics [11, 83, 85]. For instance, DeepMind's Gato (2022) was a pioneering general-purpose agent capable of performing hundreds of tasks, from image captioning and game playing to robotic manipulation, all using a single transformer-based model [86]. This approach expanded large-scale language model methods to multimodal robot control, demonstrating significant progress in generalizable artificial intelligence. In 2022, the Robotics at Google team introduced Robotics Transformer 1 (RT-1), trained on a real-world robot dataset containing 130,000 episodes and over 700 tasks (Figure 6) [15]. This model displayed high success rates even with unfamiliar objects and significantly outperformed conventional methods in terms of generalization capabilities. DeepMind's RoboCat (2023), a self-improving robotic agent, drew attention for its ability to transfer learned knowledge across robotic arms with different morphologies [87]. Initially, RoboCat learned diverse manipulation tasks using multiple robots, rapidly acquiring new tasks from as few as around 100 demonstrations, subsequently improving its performance by generating and learning from its trials. This achievement highlights the potential of foundation models and self-supervised learning to expand robot skill sets autonomously. Leveraging foundation models and teleoperation, robots can also handle complex, soft-shaped objects such as clothing through verbal instructions (Figure 6) [88]. Alongside academic publications [89, 90, 91], startups have actively engaged in related R&D and press releases (e.g., Physical Intelligence, Figure, Sklid AI, Unitree Robotics, Agility Robotics, 1X Technologies, and Insilico Medicine).

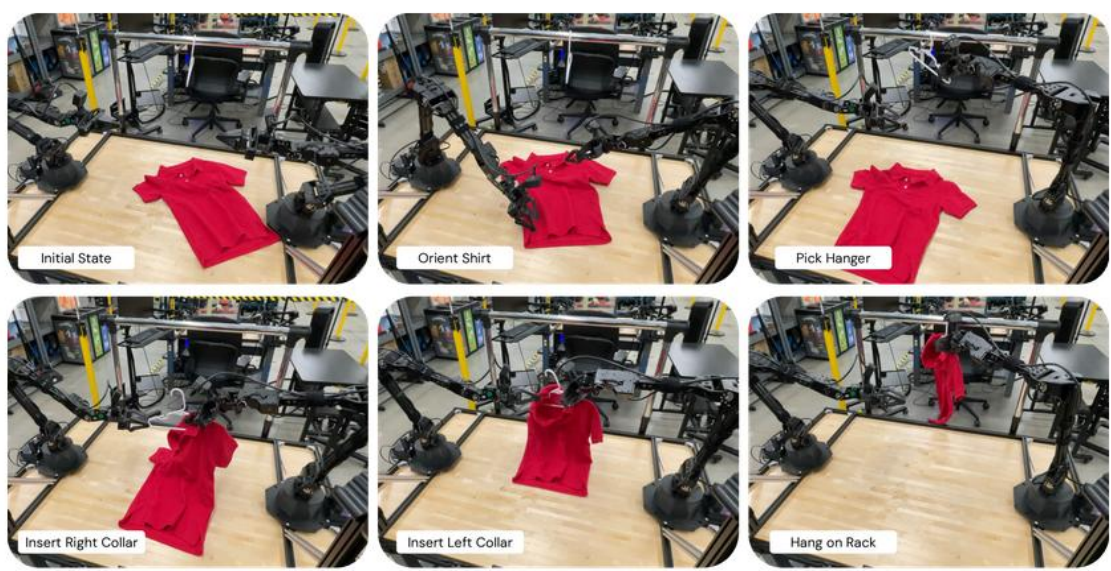

**Figure 6** Photo of a robotic foundation model system hanging a shirt on the rack. The image was reprinted from a reference under CC-BY-4.0 license [88].

**4.3 Automation in Cooking**

Efforts to apply insights from foundation models accumulated in AI and robotics fields to practical laboratory automation are beginning to emerge, although such applications remain nascent as of early 2025 [92]. Proposals include generating control programs for general-purpose robotic arms using GPT-based models [13, 93] and employing LLMs to facilitate item movements in chemical experiments [92]. However, current robotic arms and LLMs still fall short in terms of precision or response speeds required for delicate tasks, such as precise reagent additions, extractions, filtration, or sophisticated manipulation of glassware and laboratory equipment [1]. Notably, typical robot control research using foundation models often publishes demonstration videos at accelerated speeds, highlighting the unresolved issue of slow operation speeds unsuitable for experimental scenarios demanding agility [15, 51].

Cooking, which shares many characteristics with scientific experimentation, has seen more advanced foundational models and robotics research. Cooking robots must master various skills, from intricate tasks like cutting with knives or flipping pans to simpler actions such as grasping, mixing, and plating [94, 95]. Ingredients in cooking continuously change in shape and state (raw, boiled, chopped), presenting diverse challenges for robotic manipulation [95]. Furthermore, cooking requires prolonged sequences of error-free operation, including utensil manipulation and appliance usage. These complexities make cooking a representative challenge in robotics, requiring sophisticated task planning and high-precision manipulation. The soft and variable shapes of food ingredients necessitate robust design considerations, similar to those required when handling soft matter materials like polymers (e.g., coating shrimp with batter and frying it, Figure 7) [20, 96, 97].

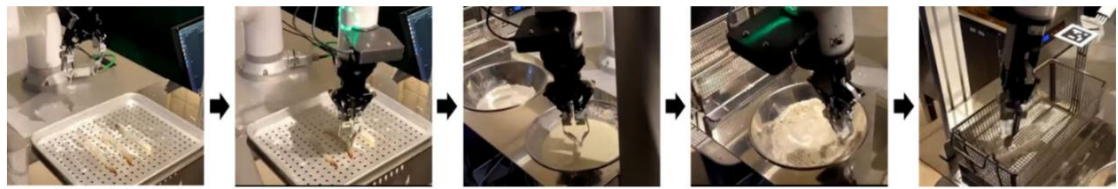

**Figure 7** A sequence of a robot system cooking shrimp fry. The image was reprinted from a reference under CC-BY-NC-4.0 license [96].

The ideal autonomous cooking system, capable of skillfully utilizing both foundation models and robots in the culinary domain, has not yet been fully realized due to software and hardware technical challenges; thus, this discussion introduces partial progress. Focusing on foundation models, there is active research on multimodal AI capable of recognizing cooking processes in real-time and dynamically instructing users. Systems integrating images, audio, and text to recognize ingredients and understand procedures have been proposed [98]. Based on such concepts, ChattyChef employs large language models to build conversational assistants that effectively guide users through cooking steps [99]. Such AI-based navigation systems could serve as transitional technologies until robotic arm technology matures.

Research aimed at teaching human culinary skills to AI and robots is also advancing. For example, studies have trained AI by showing cooking videos. Robots that memorized eight salad recipes analyzed human cooking videos, identified which recipes were being prepared, and replicated them. These robots accurately identified ingredients and actions from video footage, reproducing recipes with 93% accuracy and even creating new recipes by combining learned knowledge [100]. Such methodologies are also applicable to laboratory automation, potentially enabling AI trained on experimental procedure videos to generate new experimental protocols.

Though still imperfect, studies combining general-purpose robotic arms with foundation models for actual cooking have begun to emerge [101, 102]. Specifically, the dual-arm robot PR2 successfully executed a novel cooking recipe from scratch, following plans generated by the proposed methods [102]. For instance, it autonomously carried out tasks such as cutting, boiling, and stir-frying broccoli, guided by step-by-step instructions generated by an LLM. Google's research demonstrated a mobile robot equipped with an arm performing cooking preparation tasks like fetching ingredients from a refrigerator and heating them in a microwave, guided by LLM-generated plans [14]. Additionally, developments involving lightweight AI have enabled agile movements, successfully allowing general-purpose robotic hands to grasp soft objects rapidly [103]. Given that current robotic hands are less dexterous compared to human hands, an approach involving interchangeable specialized end-effectors for different cooking tasks has been proposed (Figure 8) [104]. This concept could similarly be applied to laboratory automation, which often requires equally complex manual skills.

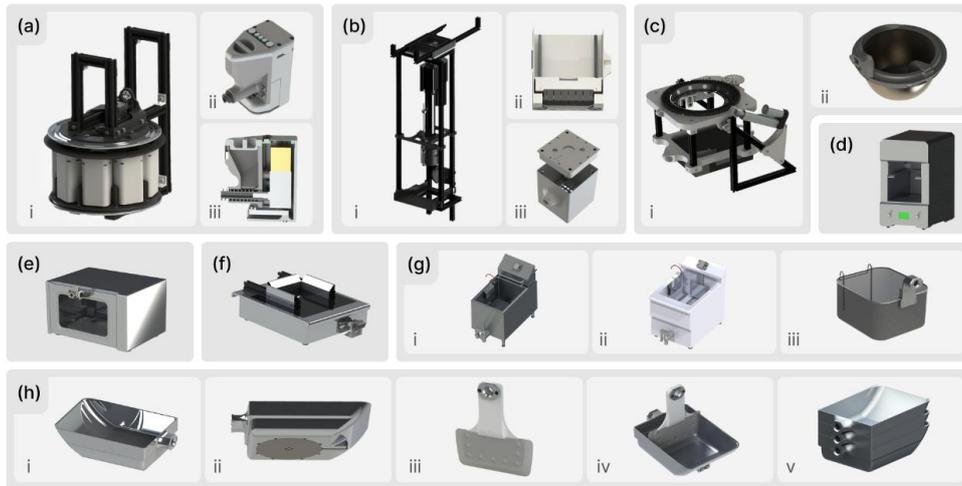

**Figure 8** Set of tools for cooking. a) Spice dispensers, b) food processor and dicing chamber, c) rotating mixer and detachable pot, d) induction pan and matching squeegee, d) convection oven, e) salamander broiler, f) induction cooktop, g) deep fryer and water boiler with their associated cooking basket, and h) custom designed induction pan and its associated squeegee sweeping tool. The images are reprinted from a reference under CC-BY-4.0 license [104].

**4.4 Cutting-edge Robotics Hardware: How Closely Can It Mimic Human Movements?**

As implied by studies in culinary applications, robotics in laboratory automation significantly relies not only on foundation models but also on high-performance hardware [1]. One of the most challenging aspects of robots performing diverse tasks in laboratories is achieving manipulation capabilities that closely approach human dexterity. As of 2025, most commercially available robotic hands feature actuators resembling tongs with two rigid moving parts rather than five-finger designs [1, 105, 106]. Such simplified mechanisms are driven by constraints such as manufacturing costs and the complexity of AI-based control systems [1]. While current hardware and its derivatives can manage certain experimental tasks, more advanced actuators are required to fully substitute human labor [1, 105, 106]. Recent research has actively explored the development of five-finger robotic hands and soft actuators aiming at delicate force control and adaptability to objects with complex geometries [107, 108, 109].

In soft robotics, technologies mimicking biological muscles and tendons through flexible materials have garnered attention, particularly for handling irregularly shaped or fragile samples previously challenging for rigid structures [109, 110, 111, 112]. Progress in soft actuators and tactile feedback technology has been notable, expanding the possibilities for safely handling glassware and samples wetted with organic solvents, and achieving high-precision pipetting [113]. For instance, a soft robotic hand has successfully executed tasks like gripping a teapot and carefully pouring liquid (Figure 9) [110]. Nevertheless, these actuator technologies still significantly lag behind human muscular and tactile capabilities in terms of speed and force, necessitating continued research and development [106, 110, 113].

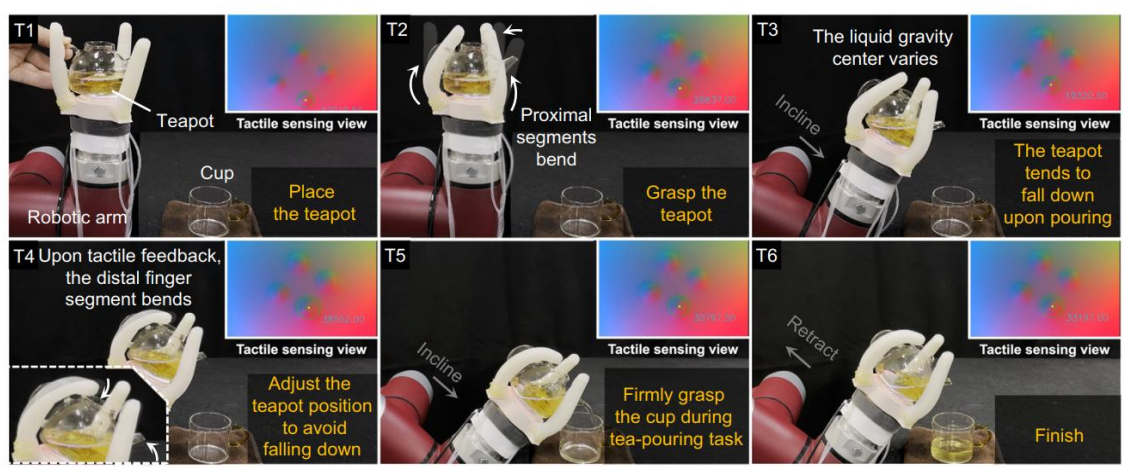

**Figure 9** A soft robotic hand grasping a teapot to pour tea. The images are reprinted from a reference under the CC-BY-NC 4.0 license [110].

Laboratory automation could take entirely different directions from human approaches. For instance, there are reports on ultrasonic technologies that control droplets in mid-air to carry out minute chemical reactions [114]. Additionally, research on electronically manipulating insect movements suggests possibilities for operations in micro-scale environments, which is challenging for conventional electronic hardware [115]. In any case, comprehensive evaluations of durability, maintainability, and cost are essential before practical laboratory automation implementation. Laboratories handling chemical experiments and biological samples require special care, as contamination or degradation due to solvents can significantly increase the risk of robot failures.

## 5. Future Roadmap and Conclusion

The previous sections reviewed the recent significant advancements linking foundation models, robotics, and laboratory automation. This section presents a roadmap for next-generation laboratory automation based on the insights gathered from this literature survey while also addressing key challenges for successful implementation. The development of foundation model-based laboratory automation can be examined along two axes: autonomy of decision-making mechanisms (analogous to the "brain") and automation of operational mechanisms (analogous to the "body"; Figure 10) [1].

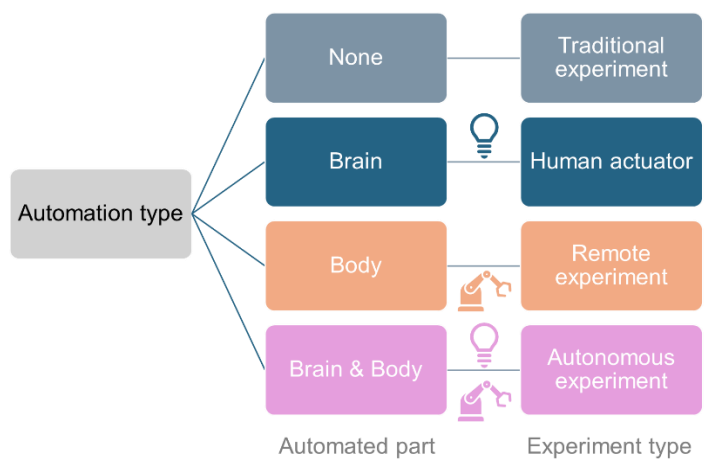

**Figure 10** A scheme for analyzing laboratory automation approaches along two axes: "brain" and "body".

Autonomous decision-making involves AI handling decisions and planning, including experimental design, equipment selection, and procedural instructions. By introducing foundation models into current laboratory automation we can expect significant advances in areas like device orchestration, data transformation, experimental planning optimization, and hardware control programming. Rapid developments in agent-based models suggest they will play important roles within several years [57, 59, 61, 62]. However, technical challenges persist, such as handling multimodal experimental data, integrating deep scientific insights into planning, and ensuring reliable control systems, all of which may require more time to address. Providing openly licensed datasets and benchmarks can help accelerate solutions.

Automation of operational mechanisms refers to the execution of physical experiments using dedicated hardware, general-purpose robots, or human collaboration. Automation through specialized hardware extends current systems and, when guided by foundation models, promises enhanced performance. General-purpose robots represent an ideal final form of automation but currently face limitations due to robotic hand constraints, making complete replication of human dexterity challenging. Thus, partial automation through specialized attachments or human-assisted remote control is currently more practical.

Another viable approach involves humans performing experiments guided by detailed AI instructions. While this differs from strict laboratory automation, it offers a practical solution for standardizing complex tasks. In many corporate settings, experimenters are required to follow supervisors' directions meticulously; however, variability and personnel costs pose challenges. The "human actuator" model—where humans execute detailed AI instructions—is economically and operationally reasonable. While raising ethical concerns, positively viewed, AI could learn skilled human practices, aiding training and skill transfer. Multimodal process data collected in this manner could become valuable assets for robotic automation development [101, 107].

The parallel advancement of decision-making sophistication and mechanical automation is essential for achieving the ultimate goal of fully autonomous experiments Clearly defining specific use cases, identifying software and hardware challenges, and iteratively refining solutions will be critical moving forward.

In conclusion, as highlighted in this study, AI research advancements are anticipated to increasingly enable digital agents for practical applications [57, 59, 61, 62]. However, conditions for effective data collection and usage in real-world laboratory automation remain insufficient, creating a bottleneck. As a response, identifying, creating, and benchmarking datasets for diverse tasks performed by domain-specific researchers are critical [116]. Publishing laboratory automation challenges as databases or benchmarks can clarify needs and data availability, encouraging participation from AI and robotics researchers [116]. Successfully executing this improvement cycle requires expanding the talent pool capable of bridging experimentation, AI, and robotics, necessitating strategic educational programs.

## Acknowledgement

This work was partially supported by a Grant-in-Aid for Scientific Research (No. 25K08763) from the Ministry of Education, Culture, Sports, Science, and Technology, Japan, and by the JST FOREST Program (Grant Number JPMJFR213V). The manuscript was originally drafted by the authors in Japanese, translated into English using GPT-4.5, and then carefully revised by the authors. We are grateful for the valuable scientific discussions with Prof. Taro Hitosugi (The University of Tokyo).

## Disclosure statement
The authors declare no conflict of interest.